\documentclass{article}

\usepackage{microtype}
\usepackage{graphicx}
\usepackage{subfigure}
\usepackage{booktabs} 

\usepackage{hyperref}
\usepackage{breakurl}

\usepackage[accepted]{icml2018}

\icmltitlerunning{NAPS: Natural Program Synthesis Dataset}

\begin{document}

\twocolumn[
\icmltitle{NAPS: Natural Program Synthesis Dataset}



\icmlsetsymbol{equal}{*}

\begin{icmlauthorlist}
\icmlauthor{Maksym Zavershynskyi}{near}
\icmlauthor{Alex Skidanov}{near}
\icmlauthor{Illia Polosukhin}{near}
\end{icmlauthorlist}

\icmlaffiliation{near}{NEAR}

\icmlcorrespondingauthor{Maksym Zavershynskyi}{max@near.ai}

\icmlkeywords{Machine Learning, ICML}

\vskip 0.3in
]



\printAffiliationsAndNotice{}  

\newcommand{\dataseturl}[0]{\url{https://near.ai/research/naps/}}
\newcommand{\datasetatrain}[0]{16410}
\newcommand{\datasetbtrain}[0]{2231}
\newcommand{\datasetbtrainpartial}[0]{1649}
\newcommand{\datasettest}[0]{485}
\newcommand{\bestresult}[0]{8.8}
\newcommand{\pseudocodesperexample}[0]{300}

\begin{abstract}
We present a program synthesis-oriented dataset consisting of human written problem statements and solutions for these problems. The problem statements were collected via crowdsourcing and the program solutions were extracted from human-written solutions in programming competitions, accompanied by input/output examples. We propose using this dataset for the program synthesis tasks aimed for working with real user-generated data. 
As a baseline we present few models, with the best model achieving \bestresult{}\% accuracy, showcasing both the complexity of the dataset and large room for future research.
\end{abstract}

\section{Introduction}

The task of \emph{program synthesis} is to automatically find a program that satisfies user's specification.
It is a problem that has been studied since the earliest days of artificial intelligence~\cite{waldinger1969,manna1975}.
With the renewed popularity of neural networks for machine learning in recent years, neural approaches to program synthesis have correspondingly attracted greater attention from the research community, which lead to great interest in datasets for program synthesis.

Most of the recent work in the field has been focused on program synthesis from examples for single domain of programming: string transformations (RobustFill \citep{Devlin2017RobustFillNP}, Neuro-Symbolic Program Synthesis \citep{DBLP:journals/corr/ParisottoMSLZK16} and Deep API Programmer \citep{Bhupatiraju2017DeepAP}) or Karel (\citet{devlin2017neural}, \citet{bunel2018leveraging}). A more domain agnostic dataset was presented in DeepCoder \citep{DBLP:journals/corr/BalogGBNT16} but still featured very small programs. All of these results have crucial limitation that datasets were synthetically generated (with exceptions for small private test sets).

Recent examples of crowdsourced natural language to program datasets are WikiSQL \citep{2017arXiv170900103Z} and NL2Bash \citep{LinWPVZE2017:TR}.  Both of these datasets are also domain specific (with WikiSQL featuring only very simple version of SQL) and don't have programming concepts like variables and control flow. Django dataset \citep{Oda2015LearningTG} has very limited scope and each textual description is associated with one line of code. 

Worth mentioning fully natural dataset from Magic The Gathering \citep{DBLP:journals/corr/LingGHKSWB16}, that has natural language from people describing actions of cards and Java programs that perform this actions in the Magic environment. This dataset has very limited scope of programs, mostly requiring to figure out complex API of the environment.

Related field to program synthesis from natural language is semantic parsing: mapping of natural language into formal representation, which can be considered as simple programs. Recent examples of such datasets are WebQuestions \citep{Berant2013SemanticPO}, Overnight \citep{Wang2015BuildingAS}, IFTTT \citep{beltagy:acl16}. All of these datasets are limited to a specific sub-domain and a limited set of functional intents.

Additionally, there is work on latent program induction which does not require programs as supervision. This simplifies the dataset collection but has a limitation that programs frequently fail to generalize to different inputs \citep{Graves2014NeuralTM} and does not expose interpretable program back to the user while having huge performance overhead at runtime (\citet{kaiser2015neural}, \citet{neelakantan2016learning}).

In this work we presenting Natural Program Synthesis Dataset v1.0 (NAPS), freely available at \url{https://near.ai/research/naps/}, consisting of real expert programmers' solutions for complex problems and rewritten statements in the form that is approachable at the current state of technology. Dataset contains \datasetbtrain{} training and \datasettest{} test examples, with additional \datasetatrain{} unlabelled examples for pretraining and data augmentation.

\begin{table*}[t]
\caption{NAPS Dataset Structure}
\label{dataset-structure-table}
\vskip 0.15in
\begin{center}
\begin{small}
\begin{tabular}{llccc}
\toprule
Field & Description & Training A & Training B & Test \\
\midrule
Solutions & Full programs in UAST format solving a competitive problem & $\surd$ & $\surd$ & $\surd$ \\
Partial solutions & Smaller pieces of the full programs & $\times$ & $\surd$ & $\times$ \\
IO examples & Input/output examples for the full programs & $\surd$ & $\surd$ & $\surd$ \\
IO schema & Input/output types and argument names for the full programs & $\surd$ & $\surd$ & $\surd$ \\
Statements & Crowdsourced problem statements in the imperative format & $\times$ & $\surd$ & $\surd$ \\
URLs & URLs to the original problem statements & $\surd$ & $\surd$ & $\surd$ \\
\bottomrule
\end{tabular}
\end{small}
\end{center}
\vskip -0.1in
\end{table*}

\begin{table*}[t]
\caption{NAPS Dataset Metrics}
\label{dataset-metrics-table}
\vskip 0.15in
\begin{center}
\begin{small}
\begin{tabular}{lccc}
\toprule
Metric  & Training A & Training B & Test \\
\midrule
Number of examples in the dataset & \datasetatrain{} & \datasetbtrain{} & \datasettest{} \\
Number of examples that are partial solutions & --- & \datasetbtrainpartial{} & --- \\
Number of synthetic statements per solution & \pseudocodesperexample{} & --- & --- \\
\midrule
Statements length, i.e. number of tokens & 173 $\pm$ 113 (synthetic) & \multicolumn{2}{c}{93 $\pm$ 51 (real)} \\
Number of lines of code per solution & \multicolumn{3}{c}{21.7 $\pm$ 6.4} \\
Number of inputs/outputs per solution & \multicolumn{3}{c}{7.5 $\pm$ 2}\\
\bottomrule
\end{tabular}
\end{small}
\end{center}
\vskip -0.1in
\end{table*}

To assess the difficulty of the NAPS dataset, we implemented sequence-to-sequence and sequence-to-tree baselines. Our best model achieves the accuracy of \bestresult{}\%. This shows there is a lot of room for advancement both in modeling and in data augmentation on the NAPS dataset.

\section{Dataset}

The first release of the NAPS dataset is split into three portions. The largest dataset contains \datasetatrain{} competitive problem solutions with the corresponding input/output examples and URL links to the original problem statements from the codeforces.com website from which the problem statements can be retrieved. We also accompany each solution with \pseudocodesperexample{} synthetic problem statements that we used for training the baseline models, see Section~\ref{experimental-results}.

The second dataset contains \datasetbtrain{} competitive programming solutions together with the partial exerts from problem solutions. Each record in this dataset is accompanied with a problem statement that was collected by the means of a crowdsourcing platform, a URL to the original problem statement, and input/output examples for non-partial solutions.

The third, smallest dataset contains \datasettest{} full problem solutions also accompanied with the crowdsourced problem statements, URLs, and input/output examples.

\textbf{Solutions:} The solutions presented in this dataset are collected from the programming competitions. We then have converted the code written in Java into our intermediate language, UAST, which additionally allowed us to unify library-specific containers and algorithms. In the future, this method will also allow our models to work with solutions across programming languages such as C++, Python, C\#, and Pascal.

\textbf{Written Statements:} We hosted a crowdsourcing platform with participants from competitive programming community and asked them to describe the problem solution that was presented to them in UAST. The process was moderated and the participants were strongly encouraged to give descriptions that were as high-level as possible while at the same time using the language with the imperative structure of the sentences. To provide a curriculum step for the models trained on this dataset, we also asked the participants to describe the smaller inner blocks of the solutions. The workers were allowed to reuse the language used for the inner blocks when describing the blocks enclosing them, but only if the larger block couldn't be described at a higher abstraction level.

\textbf{Tests:} Each full solution is accompanied with 2-10 inputs/outputs each split into two groups. The first group can be used in search or can be included into the problem specification as part of the model input. The second group can be used for the evaluation at the test time.

\subsection{UAST Specification}
UAST eliminates the burden of managing a runtime or having a compilation step. The code is convertible back and forth between UAST and Java while preserving the readability and the ability to run the input/output examples. While converting to UAST, we also remove all the file I/O and pass all the input data as arguments to the main function, and make the function return the final output. The execution engine and tools for static and runtime analysis can be found at \dataseturl{}.

\begin{table*}[t]
\caption{UAST Specification}
\label{uast-spec-table}
\vskip 0.15in
\begin{center}
\begin{small}
\begin{tabular}{rll}
\toprule
\texttt{PROGRAM} & \texttt{::=} & \texttt{\{'types': [RECORD...], 'funcs': [FUNC...]\}} \\
& & \textit{Optional record } \texttt{'\_\_globals\_\_'}\ \textit{declares global variables. Function} \texttt{'\_\_main\_\_'}\ \textit{is the entry point and} \\
& & \textit{optional function}\ \texttt{'\_\_globals\_\_.\_\_init\_\_'}\ \textit{initializes the global variables.} \\
\texttt{RECORD} & \texttt{::=} & \texttt{['record', name, \{field\_name: VAR...\}]} \\
\texttt{FUNC} & \texttt{::=} & \texttt{['func' \textbar\ 'ctor', TYPE, name, [VAR...], [VAR...], [STMT...]]} \\
& & \textit{The function entries are: the return type, the name, the arguments, the local variables, and the body.} \\
\texttt{VAR} & \texttt{::=} & \texttt{['var', TYPE, name]} \\
\texttt{STMT} & \texttt{::=} & \texttt{EXPR \textbar\ IF \textbar\ FOREACH \textbar\ WHILE \textbar\ BREAK \textbar\ CONTINUE \textbar\ RETURN \textbar\ NOOP} \\
\texttt{EXPR} & \texttt{::=} & \texttt{ASSIGN \textbar\ VAR \textbar\ FIELD \textbar\ CONSTANT \textbar\ INVOKE \textbar\ TERNARY \textbar\ CAST} \\
\texttt{ASSIGN} & \texttt{::=} & \texttt{['assign', TYPE, LHS, EXPR]} \\
\texttt{LHS} & \texttt{::=} & \texttt{VAR \textbar\ FIELD \textbar\ INVOKE} \\
\texttt{IF} & \texttt{::=} & \texttt{['if', TYPE, EXPR, [STMT...], [STMT...]]} \\
\texttt{FOREACH} & \texttt{::=} & \texttt{['foreach', TYPE, VAR, EXPR, [STMT...]]} \\
\texttt{WHILE} & \texttt{::=} & \texttt{['while', TYPE, EXPR, [STMT...], [STMT...]]} \\
\texttt{BREAK} & \texttt{::=} & \texttt{['break', TYPE]} \\
\texttt{CONTINUE} & \texttt{::=} & \texttt{['continue', TYPE]} \\
\texttt{RETURN} & \texttt{::=} & \texttt{['return', TYPE, EXPR]} \\
\texttt{NOOP} & \texttt{::=} & \texttt{['noop']} \\
\texttt{FIELD} & \texttt{::=} & \texttt{['field', TYPE, EXPR, field\_name]} \\
\texttt{CONSTANT} & \texttt{::=} & \texttt{['val', TYPE, value]} \\
\texttt{INVOKE} & \texttt{::=} & \texttt{['invoke', TYPE, function\_name, [EXPR...]]} \\
\texttt{TERNARY} & \texttt{::=} & \texttt{['?:', TYPE, EXPR, EXPR, EXPR]} \\
\texttt{CAST} & \texttt{::=} & \texttt{['cast', TYPE, EXPR]} \\
\texttt{TYPE} & \texttt{::=} & \texttt{bool \textbar\ char \textbar\ int \textbar\ real \textbar\ TYPE* \textbar\ TYPE\% \textbar\ \textless TYPE\textbar TYPE\textgreater\ \textbar\ record\_name\#} \\
& & \textit{The last four types correspond to an array, a set, a map, and a record type.} \\
\bottomrule
\end{tabular}
\end{small}
\end{center}
\vskip -0.1in
\end{table*}

The language allows several redundancies that simplify the code analysis and the implementation of the executor and the tools. For instance, each expression has a \texttt{TYPE} as the second entry which eliminates the need for deducing the types. Functions require declaring local variables in advance, see Table~\ref{uast-spec-table}. We have also introduced \texttt{FOREACH} and \texttt{TERNARY} which can be expressed through other control-flow constructs but their introduction has greatly reduced the size of the code. In addition the language is accompanied with a short library of basic functions like \texttt{'map\_keys'}, \texttt{'string\_find'}, etc.

\section{Experimental Results}~\label{experimental-results}
In this section, we present some of our results on applying sequence-to-sequence and sequence-to-tree models for synthesizing programs from problem statements. In addition, we present the data-structure that we used to perform the decoding in the sequence-to-tree model.

At training time we construct batches by first choosing between datasets A and B with equal probability and then drawing a random example from the chosen dataset. For sequence-to-tree model we choose dataset A 3.3-times more frequently than B. For the dataset A we generated synthetic problem statements using a rule-based randomized method where the rules were selected to match the stylistics of the crowdsource workers as close as possible. The synthetic statements were regenerated anew at the beginning of each epoch and we include \pseudocodesperexample{} synthetic statements for each solution in the dataset A which corresponds to the number of epochs we trained our baseline models for. The evaluation was performed on the holdout dataset that did not share solutions with the training datasets.

Our sequence-to-sequence model consists of the text encoder and the program decoder mediated through the standard multiplicative attention mechanism ~\citep{DBLP:journals/corr/LuongPM15}. The encoder is the bidirectional RNN with GRU cells stacked in two layers ~\citep{DBLP:journals/corr/ChoMGBSB14}. The decoder is a single RNN with GRU cells augmented with a pointer mechanism ~\citep{2015arXiv150603134V}. In addition to using the pointer mechanism for copying out-of-vocabulary constants and string literals from problem statements to the synthesized code, we also use it for copying in-vocabulary tokens like arithmetic operations and variable names. For this reason, we preferred the soft-switch design described in~\citet{DBLP:journals/corr/SeeLM17}, which is suitable for in-vocabulary copying, over the hard-switch design described in~\citet{DBLP:journals/corr/GulcehreANZB16}.

\subsection{Sequence to Tree}
The sequence-to-tree model shares the same encoder and the attention mechanism with the sequence-to-sequence model but the decoding step accounts for the hierarchical nature of the program which allows us to restrict the output of the decoder based on the language grammar. It is done by first implementing a general purpose persistent tree data-structure~\citep{Sarnak:1986:PPL:6138.6151} that allows storing and extending multiple UASTs simultaneously, similarly to how it is done in~\citet{DBLP:journals/corr/abs-1802-04335}. The data-structure and the specific implementation of the decoder then work together where the decoder provides the nodes to extend and the data-structure extends them by forking a new tree and placing it in the priority queue based on the tree's priority, e.g. the likelihood of the entire tree defined by the logits returned from the decoder.

Each node in UAST has an access to its siblings and the parent. For each tree, we store the global state of the entire tree and for each node, we store two states: for its siblings and for its children. The data-structure then passes these states to the decoder which decides which of the incomplete nodes to extend based on the given states. The data-structure then handles multiple extension options for each node which is used in the search. In this paper, we only provide results for the decoder that always extends the left-most incomplete node based on the global state of the tree. However this design can also easily adopt decoders from other papers, e.g. \citet{DBLP:journals/corr/abs-1802-04335} and \citet{DBLP:journals/corr/ParisottoMSLZK16}.

Dataset B and the Test dataset contain problem statements written by real users which poses a challenge since personal writing style varies a lot even though we tried to incentivize the consistency. The biggest challenge is the variance in the verbosity and the usage of rare words. Rules for the synthetic problem statements attempt to mimic the variance in the style but nevertheless, the resulting model is still very sensitive to verbosity. Specifically, the model learns to assign a higher significance to out-of-vocabulary tokens during training than what is optimal for the test dataset.

For the sequence-to-sequence model, the evaluation was performed using the beam search with the beam size equal 64. For the sequence-to-tree model the queue capacity was 64 and at each step, the decoder would expand the left-most incomplete node with 64 most probable tokens yielding 64 new trees which would utilize the memory saving properties of the persistent trees. In the end, we would search through the resulting 64 programs and pick the one that passed the input/output tests. The accuracy is then measured by counting the synthesized programs that pass all the input/output tests that were not used in the search. We also define 50\%accuracy metric which counts the programs that pass at least 50\% of the test input/output examples.

\begin{table}[t]
\caption{Accuracy of vanilla and pointer models with and without out-of-vocabulary copying}
\label{accuracy-table}
\vskip 0.15in
\begin{center}
\begin{small}
\begin{sc}
\begin{tabular}{lcc}
\toprule
Model & Accuracy & 50\%Accuracy\\
\midrule
Vanilla Seq2Seq & 0\% & 0\% \\
Seq2Seq wihout OOV & 3.5\% & 5.9\% \\
Seq2Seq with OOV & 4.7\% & 7\%\\
Seq2Tree without OOV & 8.8\% & 11.2\%\\
Seq2Tree with OOV & 7.9\% & 10.2\%\\
\bottomrule
\end{tabular}
\end{sc}
\end{small}
\end{center}
\vskip -0.1in
\end{table}

\begin{table}[t]
\caption{Example of the inferred program and the tests}
\label{inf-example}
\vskip 0.15in
\begin{center}
\begin{small}
\begin{tabular}{rrrrrrrr}
\toprule
\multicolumn{8}{l}{int \_\_main\_\_(int var0)} \\
\multicolumn{8}{l}{\quad vars: int var1,  int var2,  int var3} \\
\multicolumn{8}{l}{\quad var2 = 2} \\
\multicolumn{8}{l}{\quad if (((var0 - 2) \% 3) == 0)} \\
\multicolumn{8}{l}{\qquad var1 = 1} \\
\multicolumn{8}{l}{\quad else} \\
\multicolumn{8}{l}{\qquad var1 = 0} \\
\multicolumn{8}{l}{\quad var3 = 1} \\
\multicolumn{8}{l}{\quad for(; (var3 $<$ var0); var3 = (var3 + 1))} \\
\multicolumn{8}{l}{\qquad if (var2 $<$ var0)} \\
\multicolumn{8}{l}{\qquad\quad var2 = (var2 + ((var3 * 3) + 2))} \\
\multicolumn{8}{l}{\qquad\quad if (((var0 - var2) $\geq$ 0) \& ((var0 - var2) $\leq$ 0))} \\
\multicolumn{8}{l}{\qquad\qquad var1 = (var1 + 1)} \\
\multicolumn{8}{l}{\qquad\quad else} \\
\multicolumn{8}{l}{\qquad\qquad if (((var0 - var2) $\geq$ 0) \& (((var0 - var2) \% 3) == 0))} \\
\multicolumn{8}{l}{\qquad\qquad\quad var1 = (var1 + 1)} \\
\multicolumn{8}{l}{\qquad else} \\
\multicolumn{8}{l}{\qquad\quad break} \\
\multicolumn{8}{l}{\quad return var1} \\
\midrule
Search Input & 157 & 1312861 & 6  & & & &\\
Search Output & 3 &  312     & 0  & & & &\\
\midrule
Test Input & 26 & 152 & 158 & 4 & 71 & 3 & 155 \\
Test Output & 2  & 3   & 4   & 0 & 2  & 0 & 4 \\
\bottomrule
\end{tabular}
\end{small}
\end{center}
\vskip -0.1in
\end{table}

Interestingly, even when the model does mistakes during the inference those mistakes might be benign and it will still be passing the tests. For instance, Table~\ref{inf-example} shows the inference example for the following problem statement:

\textit{You are given a number var0. You have to set var2 to 2. If var0-2 is divisible by 3 you have to set var1 to 1, otherwise you have to set var1 to zero. For each var3 between 1 and var0-1, if var2 is less than var0 you have to, add var3*3+2 to var2, if var0-var2 is greater than or equal to zero and var0-var2 is divisible by 3 add 1 to var1; otherwise you have to break from the enclosing loop. You have to return var1.}

Note that if var0-var2 $\geq$ 0 \& var0-var2 $\leq$ 0 then var0-var2 $\geq$ 0 \& (var0-var2)\% 3 == 0. Even though the model has inferred a redundant if-clause it did not break the program's logic.

\section{Future Work}
NAPS dataset enables the program synthesis research on real-life non-trivial programs and problem statements written in a general-purpose language. The baseline metrics, however, demonstrate a large room for the improvement. 

\nocite{langley00}

\bibliography{example_paper}
\bibliographystyle{icml2018}

\clearpage
\onecolumn
\appendix
\section*{Appendix A. Correctly inferred programs}
This section demonstrates the examples of the programs inferred by the model and passing all input/output tests. We also provide the texts describing the problem statements and the original human-written programs. We, however, do not give the input/output tests that were used to do the search and evaluate the inferred program because their size is substantial especially for the problems that involve arrays. Please refer to the linked dataset for input/output examples.

\textbf{Example 1}
\textit{Given integers var0 , var1. Let var2 be the less of var0 and var1. Assign ( var0 + var1 ) / 3 to var3. Return the less of var2 and var3.}

\begin{table}[h]
\vskip 0.15in
\begin{center}
\begin{small}
\parbox{.45\linewidth}{
\begin{tabular}{l}
Golden code \\
\toprule
int \_\_main\_\_(int var0, int var1) \\
\quad vars: int var2,  int var3 \\
\quad if (var0 $>$ var1) \\
\qquad var2 = var1 \\
\quad else \\
\qquad var2 = var0 \\
\quad var3 = (var0 + var1) / 3 \\
\quad return (var2 $>$ var3)?var3:var2 \\
\bottomrule
\end{tabular}
}
\hfill
\parbox{.45\linewidth}{
\begin{tabular}{l}
Inferred code \\
\toprule
int \_\_main\_\_(int var0, int var1) \\
\quad vars:  int var2,  int var3,  int var4 \\
\quad var2 = min(var0, var1) \\
\quad var3 = ((var0 + var1) / 3) \\
\quad return min(var2, var3) \\
\bottomrule
\end{tabular}
}
\end{small}
\end{center}
\end{table}

\textbf{Example 2}
\textit{You are given an array of numbers var2 (indexing is 0-based), an array of numbers var3 (indexing is 0-based) and a number var1. You have to set var9 to -1000000000. For each position var10 in var2 if var3[var10] is greater than var1 you have to store in var9 the maximum between var9 and var2[var10]-var3[var10]+var1, otherwise you have to store in var9 the maximum between var9 and var2[var10]. You have to return var9.}

\begin{table}[h]
\vskip 0.15in
\begin{center}
\begin{small}
\parbox{.45\linewidth}{
\begin{tabular}{l}
Golden code \\
\toprule
int func0(int var0, int var1, int* var2, int* var3, int var4, int var5) \\
\quad vars:  int var6,  int var7,  int var8,  int var9,  int var10, \\
\qquad int var11,  int var12 \\
\quad var6 = len(var2) \\
\quad var7 = var1 \\
\quad var9 = -1000000000 \\
\quad var10 = 0 \\
\quad for(; (var10 $<$ var6); var10 = (var10 + 1)) \\
\qquad var11 = var2[(var4 = (var4 + 1) - 1)] \\
\qquad var12 = var3[(var5 = (var5 + 1) - 1)] \\
\qquad if (var12 $>$ var7) \\
\qquad\quad var8 = (var11 - (var12 - var7)) \\
\qquad else \\
\qquad\quad var8 = var11 \\
\qquad var9 = max(var9, var8) \\
\quad return var9 \\

int \_\_main\_\_(int var1, int* var2, int* var3) \\
\quad vars:  int var4,  int var5 \\
\quad var5 = 0 \\
\quad var4 = 0 \\
\quad return func0(1, var1, var2, var3, var4, var5) \\
\bottomrule
\end{tabular}
}
\hfill
\parbox{.45\linewidth}{
\begin{tabular}{l}
Inferred code \\
\toprule
int func0(int var0, int var1, int* var2, int* var3, int var4) \\
\quad vars:  int var5,  int var6,  int var7,  int var8,  int var9,  int var10 \\
\quad var5 = len(var2) \\
\quad var6 = 0 \\
\quad var7 = len(var2) \\
\quad var8 = 0 \\
\quad var9 = -1000000000 \\
\quad var10 = 0 \\
\quad for(; (var10 $<$ var5); var10 = (var10 + 1)) \\
\qquad if (var3[var10] $>$ var1) \\
\qquad\quad var9 = max(var9, ((var2[var10] - var3[var10]) + var1)) \\
\qquad else \\
\qquad\quad var9 = max(var9, var2[var10]) \\
\qquad return var9 \\
int \_\_main\_\_(int var1, int* var2, int* var3) \\
\quad vars:  int var4 \\
\quad var4 = 0 \\
\quad return func0(1, var1, var2, var3, var4) \\
\bottomrule
\end{tabular}
}
\end{small}
\end{center}
\end{table}
\pagebreak
\textbf{Example 3}
\textit{You are given an integer var0. If var0 \% 10 is less than 5 then decrement var0 by the value var0 \% 10. Otherwise, increment var0 by the value 10-var0 \% 10. Return var0.}

\begin{table}[h]
\vskip 0.15in
\begin{center}
\begin{small}
\parbox{.45\linewidth}{
\begin{tabular}{l}
Golden code \\
\toprule
int \_\_main\_\_(int var0) \\
\quad vars:  int var1 \\
\quad var1 = (var0 \% 10) \\
\quad if (var1 $<$ 5) \\
\qquad var0 = (var0 - var1) \\
\quad else \\
\qquad var0 = (var0 + (10 - var1)) \\
\quad return var0 \\
\bottomrule
\end{tabular}
}
\hfill
\parbox{.45\linewidth}{
\begin{tabular}{l}
Inferred code \\
\toprule
int \_\_main\_\_(int var0) \\
\quad vars:  int var1 \\
\quad var1 = 0 \\
\quad if ((var0 \% 10) $<$ 5) \\
\qquad var0 = (var0 - (var0 \% 10)) \\
\quad else \\
\qquad var0 = (var0 + (10 - (var0 \% 10))) \\
\quad return var0 \\
\bottomrule
\end{tabular}
}
\end{small}
\end{center}
\end{table}

\textbf{Example 4}
\textit{Given integers var0, var1. Initialize var2 to 0, var3 to 1. While var3 is not less than 1, var3 is not greater than var0 and var1 is not less than var3, perform the following operations. Add var3 to var2. Subtract var3 from var1. Increase var3 by 1. If var3 is greater than var0, set var3 to 1. Return var1.}

\begin{table}[h]
\vskip 0.15in
\begin{center}
\begin{small}
\parbox{.45\linewidth}{
\begin{tabular}{l}
Golden code \\
\toprule
int \_\_main\_\_(int var0, int var1) \\
\quad vars:  int var2,  int var3 \\
\quad var2 = 0 \\
\quad var3 = 1	\\
\quad for(; (((var3 $\geq$ 1) \& (var3 $\leq$ var0)) \& (var1 $\geq$ var3)); ) \\
\qquad if (var1 $\geq$ var3) \\
\qquad\quad var2 = (var2 + var3) \\
\qquad\quad var1 = (var1 - var3) \\
\qquad\quad var3 = (var3 + 1) \\
\qquad\quad if (var3 $>$ var0) \\
\qquad\qquad var3 = 1 \\
\qquad\quad else \\
\qquad\qquad pass \\
\qquad else \\
\qquad\quad pass \\
\quad return var1 \\
\bottomrule
\end{tabular}
}
\hfill
\parbox{.45\linewidth}{
\begin{tabular}{l}
Inferred code \\
\toprule
int \_\_main\_\_(int var0, int var1) \\
\quad vars:  int var2,  int var3 \\
\quad var2 = 0 \\
\quad var3 = 1 \\
\quad for(; (((var3 $\leq$ var0) \& (var1 $\geq$ var3)) \& (var1 $\geq$ var3)); ) \\
\qquad var2 = (var2 + var3) \\
\qquad var1 = (var1 - var3) \\
\qquad var3 = (var3 + 1) \\
\qquad if (var3 $>$ var0) \\
\qquad\quad var3 = 1 \\
\qquad else \\
\qquad\quad pass \\
\quad return var1 \\
\bottomrule
\end{tabular}
}
\end{small}
\end{center}
\end{table}

\textbf{Example 5}
\textit{Given integer var0. Initialize var2 to 1. While var2 is less than var0 subtract var2 from var0 and increase var2 by 1. Return var0.}

\begin{table}[h]
\vskip 0.15in
\begin{center}
\begin{small}
\parbox{.45\linewidth}{
\begin{tabular}{l}
Golden code \\
\toprule
int \_\_main\_\_(int var0) \\
\quad vars:  int var1,  int var2 \\
\quad var1 = 0 \\
\quad var2 = 1 \\
\quad for(; True; var2 = (var2 + 1)) \\
\qquad if (var2 $\geq$ var0) \\
\qquad\quad break \\
\qquad else \\
\qquad\quad pass \\
\qquad var0 = (var0 - var2) \\
\quad return var0 \\
\bottomrule
\end{tabular}
}
\hfill
\parbox{.45\linewidth}{
\begin{tabular}{l}
Inferred code \\
\toprule
int \_\_main\_\_(int var0) \\
\quad vars:  int var1,  int var2 \\
\quad var2 = 1 \\
\quad for(; (var2 $<$ var0); ) \\
\qquad var0 = (var0 - var2) \\
\qquad var2 = (var2 + 1) \\
\quad return var0 \\
\bottomrule
\end{tabular}
}
\end{small}
\end{center}
\end{table}

\pagebreak
\section*{Appendix B. Incorrectly inferred programs}
\textbf{Example 1}
\textit{Given an integer var0, find the smallest integer, which is greater than var0, such that all its digits are distinct and return the array, which contains one element equal to the result.}

This problem provides a great example of how our model attempts to handle more complex concepts, like comparing numbers digit-wise. In this case, it attempts to extract the digit from a number by accessing it as an array, instead of doing it through the arithmetic operations like in the golden code.

\begin{table}[h]
\vskip 0.15in
\begin{center}
\begin{small}
\parbox{.45\linewidth}{
\begin{tabular}{l}
Golden code \\
\toprule
int* \_\_main\_\_(int var0) \\
\quad vars:  int var1,  int var2,  int var3,  int var4,  int* var5 \\
\quad var5 = new int*() \\
\quad var0 = (var0 + 1) \\
\quad for(; True; ) \\
\qquad  var1 = (var0 \% 10) \\
\qquad var2 = ((var0 / 10) \% 10) \\
\qquad var3 = ((var0 / 100) \% 10) \\
\qquad var4 = (var0 / 1000) \\
\qquad if ((((((var1 != var2) \& (var1 != var3)) \& (var1 != var4)) \\
\qquad\quad \& (var2 != var3)) \& (var2 != var4)) \& (var3 != var4)) \\
\qquad\quad array\_push(var5, var0) \\
\qquad\quad break \\
\qquad else \\
\qquad\quad pass \\
\qquad var0 = (var0 + 1) \\
\quad return var5 \\
\bottomrule
\end{tabular}
}
\hfill
\parbox{.45\linewidth}{
\begin{tabular}{l}
Inferred code \\
\toprule
int* \_\_main\_\_(int var0) \\
\quad vars:  int var1,  int var2,  int var3,  int* var4 \\
\quad var4 = new int*() \\
\quad var2 = 0 \\
\quad var3 = 0 \\
\quad for(; (var3 $<$ var2); var3 = (var3 + 1)) \\
\qquad var4 = var0[(var2 = (var2 + 1) - 1)] \\
\qquad if (var2[var3] == 1) \\
\qquad\quad var2 = (var2 + 1) \\
\qquad else \\
\qquad\quad pass \\
\quad return var4 \\
\bottomrule
\end{tabular}
}
\end{small}
\end{center}
\end{table}

\textbf{Example 2}
\textit{Given integer var0. Return the sum of powers of 2 from 1 to var0.}

For this problem, the inferred program uses incorrect variable names in the iterator increment and in the accumulator. Besides that, the program structure is correct.

\begin{table}[h]
\vskip 0.15in
\begin{center}
\begin{small}
\parbox{.45\linewidth}{
\begin{tabular}{l}
Golden code \\
\toprule
int \_\_main\_\_(int var0) \\
\quad vars:  int var1,  int var2 \\
\quad var1 = 0 \\
\quad var2 = 1 \\
\quad for(; (var2 $\leq$ var0); var2 = (var2 + 1)) \\
\qquad var1 = (var1 + pow(2, var2)) \\
\quad return var1 \\
\bottomrule
\end{tabular}
}
\hfill
\parbox{.45\linewidth}{
\begin{tabular}{l}
Inferred code \\
\toprule
int \_\_main\_\_(int var0) \\
\quad vars:  int var1,  int var2,  int var3 \\
\quad var1 = 1 \\
\quad var2 = 1 \\
\quad for(; (var2 $\leq$ var0); var3 = (var3 + 1)) \\
\qquad var3 = (var3 + (2 $<<$ var2)) \\
\quad return var2 \\
\bottomrule
\end{tabular}
}
\end{small}
\end{center}
\end{table}

\pagebreak
\textbf{Example 3}
\textit{You are given integers var0, var1. Return the factorial of the minimum of var1 and var0.}

This is an example of a completely incorrect inference.

\begin{table}[h]
\vskip 0.15in
\begin{center}
\begin{small}
\parbox{.45\linewidth}{
\begin{tabular}{l}
Golden code \\
\toprule
int \_\_main\_\_(int var0, int var1) \\
\quad vars:  int var2,  int var3,  int var4 \\
\quad if (var0 $>$ var1) \\
\qquad var3 = var1 \\
\quad else \\
\qquad var3 = var0 \\
\quad var4 = var3 \\
\quad for(; (var4 $>$ 1); var4 = (var4 - 1)) \\
\qquad var3 = (var3 * (var4 - 1)) \\
\quad return var3 \\
\bottomrule
\end{tabular}
}
\hfill
\parbox{.45\linewidth}{
\begin{tabular}{l}
Inferred code \\
\toprule
int \_\_main\_\_(int var0, int var1) \\
\quad vars:  int var2 \\
\quad return (min(var1, var0) * 2) \\
\bottomrule
\end{tabular}
}
\end{small}
\end{center}
\end{table}

\section*{Appendix C. Decoding algorithm}
Here we give a high-level description of how the data-structure and the decoder perform the decoding in the sequence-to-tree model.

\begin{algorithm}[h]
\begin{algorithmic}
    \STATE {$Q \leftarrow \{create\_empty\_tree()\}$}
    \WHILE{$has\_incomplete\_trees(Q)$}
        \FORALL{$T$ \textbf{in} $incomplete\_trees(Q)$}
            \STATE $candidates \leftarrow \emptyset$
            \FORALL{$N$ \textbf{in} $incomplete\_nodes(T)$}
                \STATE $h_{tree} \leftarrow T.h$
                \STATE $h_{parent} \leftarrow parent(N).h_{parent}$
                \STATE $h_{left} \leftarrow extract\_states(left\_siblings(N))$
                \STATE $h_{right} \leftarrow extract\_states(right\_siblings(N))$
                \STATE $candidates.add((h_{tree}, h_{parent}, h_{left}, h_{right}))$
            \ENDFOR
            \STATE $new\_nodes \leftarrow \textbf{DECODER}(candidates)$
            \FORALL{$N$ \textbf{in} $new\_nodes$}
                \IF{$N$ satisfies language grammar}
                \STATE {Fork a new tree $T_{new}$ by extending $N$}
                \STATE {$Q.push(T_{new})$}
                \ENDIF
            \ENDFOR
        \ENDFOR
    \ENDWHILE
\end{algorithmic}
\end{algorithm}

\end{document}